\documentclass[%
reprint,
 amsmath,amssymb,
 aps,
prx
]{revtex4-2}

\usepackage{graphicx}% Include figure files
\usepackage{dcolumn}% Align table columns on decimal point
\usepackage{bm}% bold math
\usepackage{xcolor}
\usepackage[normalem]{ulem}%\usepackage{hyperref}% add hypertext capabilities
\usepackage[mathlines]{lineno}% Enable numbering of text and display math
\begin{document}

\preprint{APS/123-QED}

\title{The Physics Is the Network: Fast, Training-Free Neural Emulation}% Force line breaks with \\
% \title{Leveraging Neural Networks to Model Chaotic }

\author{Luka Todorovic}
% 
 
\email{\href{mailto:lukat@umich.edu}{lukat@umich.edu}}
\affiliation{%
 Department of Mathematics and Department of Statistics, University of Michigan, Ann Arbor, MI
}%

% \collaboration{MUSO Collaboration}%\noaffiliation

\author{Kaushal Gianchandani}
\affiliation{
NASA Postdoctoral Program Fellow, NASA Goddard Institute for Space Studies, New York, NY
}%
% \altaffiliation[Also at ]{Physics Department, XYZ University.}

\author{Sai Ravela}
\email{\href{mailto:ravela@mit.edu}{ravela@mit.edu}}
\affiliation{%
Department of Earth, Atmospheric, and Planetary Sciences,
Massachusetts Institute of Technology, Cambridge, MA
}

% \collaboration{CLEO Collaboration}%\noaffiliation

\date{\today}% It is always \today, today,

\begin{abstract}
Neural emulation need not require learning a substitute for known physics. We investigate PolyNets that couple a continuous-time neural representation of polynomial dynamics with a simple recurrent network for discrete-time integration. When the governing coefficients are known, the resulting tensor graph requires no training and preserves the prescribed numerical scheme while enabling efficient hardware execution and parallel simulation. Using the one-dimensional Korteweg–de Vries equation, we compare PolyNet with CPU and GPU finite-difference Runge–Kutta solvers and three trained neural models: a state-map emulator, a physics-informed predictor, and a learned-tendency integrator. Analytical soliton solutions provide a reference for amplitude and phase errors across grid resolutions, \(10^{6}\) dispersively scaled time steps, ensembles of up to \(10^{4}\) trajectories, and varying initial amplitudes. PolyNet achieves accuracy comparable to the numerical solvers, including at amplitudes outside the learned models’ training conditions. It combines this fidelity with high GPU throughput, approximately constant over the tested resolution range, and favorable ensemble scaling.
% covered at length in the main body of the article. 
% \begin{description}
% Secondary publications and information retrieval purposes.
% You may use the \texttt{description} environment to structure your abstract;
% use the optional argument of the \verb+\item+ command to give the category of each item. 
% \end{description}
\end{abstract}

\maketitle

%\tableofcontents
\section{Introduction}

Learning physical dynamics and executing them through a neural network are different tasks. When governing equations are known, their mathematical structure can determine the network architecture~\cite{Trautner2019}, avoiding the need to infer from trajectories what is already prescribed by the physics. For systems with polynomial nonlinearities, this distinction is concrete: products and weighted sums can represent the governing interactions directly, while recurrent operations implement numerical time integration~\cite{Trautner2019}. Neural computation can therefore serve not only as a framework for approximating physical evolution, but also as a computational realization of the equations themselves. The resulting opportunity is to exploit neural execution without introducing a learned approximation to the dynamics.    

Neural approaches to emulation differ fundamentally in what they approximate. An emulator learns a discrete mapping between successive states \cite{qin2019data}. A neural differential-equation model~\cite{chen2019neuralordinarydifferentialequations} instead learns the temporal tendency, which is integrated to advance the solution. Physics-informed neural networks incorporate governing equations into training through differential constraints on the predicted evolution~\cite{Raissi2019}. These approaches provide flexibility, but their performance depends on both the numerical procedure and the learned representation. In an autoregressive simulation, prediction errors become inputs to subsequent predictions; accuracy over long integrations is therefore a separate requirement from accuracy on individual training examples. Likewise, reproducing trajectories within a training distribution does not establish fidelity under changes in initial conditions. Evaluating a learned simulator consequently requires considering training cost, accumulated error, and generalization alongside execution speed. 

PolyNets provide an alternative in which the governing structure is prescribed rather than learned~\cite{Trautner2019}. Trautner and Ravela (2019) constructed these networks using multiplicative nodes to represent polynomial nonlinearities and weighted additive connections to combine their contributions. When the governing coefficients are known, the network requires no training. They also developed recurrent neural circuits for Runge–Kutta and Adams–Bashforth–Moulton integration, bringing tendency evaluation and time advancement into a single neural computation. They call this neural integration.

Their experiments with the Lorenz system demonstrated correspondence between neural and conventional numerical integration, with floating-point discrepancies amplified by the chaotic dynamics. The relevant equivalence is to the numerical algorithm—not to an exact solution of the continuous equations. An equation-derived PolyNet retains the approximation inherent in the chosen numerical scheme while avoiding the additional error associated with learning the governing tendency.   

That equivalence establishes expected numerical fidelity; the physics and its numerical realization were trivial neural circuits, but not a computational advantage. Trautner and Ravela proposed that tensor-based neural execution could exploit specialized hardware, particularly for ensembles of independently evolving systems, while leaving its performance for partial differential equations untested. The practical question is whether this representation can preserve the accuracy of an established numerical scheme while improving its execution at increasing spatial resolution and ensemble size. Answering it requires comparison not only with trained neural models, but also with conventional GPU solvers: acceleration relative to a CPU implementation alone cannot establish the advantage of a particular neural architecture.  

Thus, the computational pathway involves neural integration of the differential operator and its numerical integration as a tensor graph on neural hardware, GPUs, or TPUs, at native speeds.  

Here, we examine this question using the one-dimensional Korteweg–de Vries (KdV) equation. Its balance between nonlinear advection and dispersion supports soliton solutions that provide an analytical reference for assessing errors in wave amplitude and phase. These quantities test distinct but coupled aspects of numerical fidelity. An amplitude error distorts the wave profile and, because propagation speed depends on amplitude, can also produce an accumulating position error. Meanwhile, resolving the dispersive term requires increasingly small time steps as the spatial grid is refined, raising the cost of explicit integration. KdV thus permits a controlled comparison in which numerical error can be measured directly and computational demands increased systematically. Independent soliton trajectories provide an ensemble benchmark, while variations in initial amplitude test whether trained models reproduce the governing evolution beyond the conditions represented in their training data.    

We compare six implementations: CPU and GPU finite-difference fourth-order Runge--Kutta (RK4), an equation-derived PolyNet, a learned state-map Emulator, a physics-informed state-map predictor, and a learned-tendency neural PDE model. The tests measure throughput, accuracy, ensemble scaling, and performance under changed soliton heights over prescribed fixed-step workloads.

The results show that PolyNet combines the accuracy of the corresponding finite-difference Runge–Kutta scheme with competitive GPU throughput, favorable ensemble scaling, and no training requirement. Although the emulator executes faster {at coarse grids}, PolyNet achieves higher throughput at the finest grid tested while retaining numerical fidelity over long integrations. It also maintains agreement with its numerical counterpart as the initial wave amplitude changes, whereas the trained models exhibit substantially larger errors. These findings demonstrate that neural acceleration need not depend on learning a substitute for known dynamics. For the polynomial system examined here, prescribing the governing operations within the network provides a route to efficient simulation that preserves the numerical model rather than approximating it anew.

Section~\ref{sec:related} reviews equation-derived neural integration and learned physical emulators. Section~\ref{sec:methods} presents the six implementations, followed by the KdV benchmark and numerical comparisons. The discussion and conclusions summarize the scope of the results. Appendices~\ref{app:dispersive-timestep} and~\ref{app:error-scaling} derive the time-step scaling and expected amplitude and phase error behavior.

\section{Related Work}
\label{sec:related}
 {\textbf{Equation-derived neural computation.} Trautner and Ravela~\cite{Trautner2019} showed that a polynomial vector field can be represented by a network of product nodes and additive connections whose weights are fixed by the governing coefficients. Their recurrent neural integration circuits implement explicit Runge--Kutta and Adams--Bashforth--Moulton methods and were tested on the Lorenz system. The central distinction is between exact representation of the \emph{semidiscrete polynomial tendency} and the numerical approximation introduced by spatial discretization and time stepping. The present work extends that construction to a PDE benchmark and measures execution costs, accuracy, and ensemble throughput.}

 {\textbf{Learned differential equations and physics-informed models.} Neural ODEs learn a state-dependent tendency and integrate it numerically~\cite{chen2019neuralordinarydifferentialequations}; physics-informed neural networks penalize violations of prescribed differential equations during training~\cite{Raissi2019}. These methods differ from PolyNet because their representation or coefficients are optimized from data. They are valuable when the governing dynamics are unknown or require inferred closures, but introduce training-dependent errors absent from an exactly prescribed polynomial tendency.}

 {\textbf{Learned PDE solution operators.} Fourier neural operators~\cite{Li2021FNO}, message-passing PDE solvers~\cite{Brandstetter2022MPPN}, and related neural emulators learn spatial or temporal evolution maps. Their performance can be sensitive to resolution, rollout duration, and distribution shift. These approaches are not interchangeable with the equation-derived construction: one learns an approximation to an evolution operator, while the other compiles an available numerical operator into a tensor graph. The benchmark here compares those computational choices rather than proposing a new differential equation or time integrator.}

 {\textbf{Numerical reference and error analysis.} Classical RK methods and finite-difference truncation analysis provide the appropriate baseline for evaluating a prescribed neural integrator~\cite{Butcher2008,HundsdorferVerwer2003}. The KdV soliton family provides an analytical real-line reference~\cite{ZabuskyKruskal1965,DrazinJohnson1989}. Appendices~\ref{app:dispersive-timestep} and~\ref{app:error-scaling} use these standard tools to explain the implemented dispersive step and the height-dependent error trends without treating an empirical scaling as a theorem.}

\section{Integration Methods}
\label{sec:methods}

We use two classical integration methods and four `neural' integration methods to solve the one-dimensional Korteweg-De Vries (1D KdV) equation. %and the two-dimensional Kuramoto-Sivashinsky (2D KS)  equation.
Each method is evaluated on four criteria: speed, accuracy, ensemble size, and generalization. % For the 1D KdV equation, 

{At each resolution $N=100\cdot2^n$ ($n=0,\ldots,8$), we use $\Delta x=40/N$ and $\Delta t\simeq(\Delta x)^3/6$, equivalent to the operator-normalized dispersive number $\Delta t\lVert D_3\rVert_\infty\simeq0.5$ (Appendix~\ref{app:dispersive-timestep}). The fixed-step benchmark advances $10^6$ steps at each resolution. We compare setup/training time, throughput, and maximum sampled peak amplitude and phase errors, as well as ensembles of up to $10^4$ trajectories and a height sweep $h=2,2.5,3,3.5,4$. These are fixed-step-count comparisons; the physical horizon varies with $N$. {For the learned models, training uses analytical soliton states at $h=3$; the $h\neq3$ height sweep tests withheld amplitudes within the same analytical family.}} Next, we briefly describe each integration method. 

\subsection{Classical Integration Methods}

\subsubsection{Fourth-order Runge-Kutta integration on a CPU} 
The CPU solver applies periodic centered finite differences to $u_x$ and $u_{xxx}$ and advances the resulting semidiscrete KdV system by four-stage RK4. The same spatial operators define the prescribed PolyNet tendency. The CPU implementation supplies the direct numerical reference for assessing differences caused by the tensor-graph execution.

\subsubsection{RK4 integration on a GPU}
Like the CPU implementation, the GPU implementation uses finite differences to estimate the spatial derivatives ($u_x$ and $u_{xxx}$) and RK4 to advance the solution by a time step $dt$. This provides a numerical baseline for comparing the neural integrators with the same integration scheme as the CPU executed on a GPU. 

%To compare the finite differences method with spectral methods, a GPU integration method using the Fast Fourier Transforms (FFT) to estimate the spatial derivatives is also examined. To avoid numerical instability we use Orszag's Rule for de-aliasing, masking the largest third of frequencies in the FFT derivative estimates \cite{orszag1971elimination}. 

% Much like the CPU integration, the GPU integration uses the RK4 method for time integration. The integration scheme remains consistent with that of the CPU implementation, using finite differences for the spatial derivative estimation ($u_x, u_{xxx}$) and the RK4 scheme for the evaluation of $u_t$ to advance the soliton by timestep $dt$. This provides us a similar baseline to compare the neural integrators with, but this time harnessing a GPU rather than a CPU {\color{red} KG, PLEASE READ}
 
\subsection{Neural Integrators}

\subsubsection{PolyNet} 

{PolyNet constructs the spatially discretized tendency $F_N(U)=-6U\odot D_1U-D_3U$ from fixed linear operators and componentwise multiplication. This is a degree-two polynomial in the discrete state $U$; hence its polynomial representation is exact for the prescribed semidiscrete vector field when the coefficients are known~\cite{Trautner2019}. Time advancement is the full four-stage RK4 map $U^{n+1}=\mathcal R_{\Delta t}[F_N](U^n)$, not a single composition of an integrator with one tendency evaluation. While this tensor-graph implementation requires no training, its discretization and floating-point errors remain.}

%The JAX library is utilized to carry out this method. \textcolor{red}{Check if I used TensorFlow as well, it's been a minute} {\color{blue}KG: did you figure out if TensorFlow was used? Address this.} \\ 

\subsubsection{Emulator Neural Network}

{The Emulator is a convolutional neural state-map predictor $\widehat U^{n+1}=E_\theta(\widehat U^n)$, trained on analytical solitary-wave examples with mean-squared-error loss and Adam optimization. The current implementation uses three hidden convolutional layers with 64 channels. The predicted field is fed back autoregressively. {Training uses Gaussian input noise with standard deviation $0.02$. Training pairs are generated from the analytical soliton at $h=3$; the held-out amplitudes in Fig.~\ref{fig:generalization} test generalization within the same solution family.} Unlike PolyNet, this model learns the state transition rather than prescribing the KdV tendency~\cite{Li2021FNO,Brandstetter2022MPPN}.}

\subsubsection{Physics-Informed Neural Network (PINN)}
{The PINN is a discrete-time convolutional state-map predictor, $\widehat U^{n+1}=P_\theta(\widehat U^n)$. It is trained using a one-step analytical target, a physics reference generated by a spectral RK4 KdV step, and a spatial-gradient target, with an additional eight-step autoregressive rollout loss activated after 500 epochs. Its objective is}
\begin{equation}
\bm{\mathcal L}=\bm{\mathcal L}_{\mathrm{MSE}}
+\bm{\mathcal L}_{\mathrm{Phys}}
+0.1\bm{\mathcal L}_{\mathrm{Grad}}
+\bm{\lambda}_{R}\bm{\mathcal L}_{\mathrm{Roll}},
\label{eq:pinn-loss}
\end{equation}
{where $\mathcal L_{\mathrm{MSE}}$ compares the one-step prediction with the analytical soliton, $\mathcal L_{\mathrm{Phys}}$ compares it with the spectral RK4 teacher step, and $\mathcal L_{\mathrm{Grad}}$ compares spatial derivatives of the prediction and analytical target. The rollout loss averages prediction errors over eight recursively predicted states against the analytical soliton trajectory. The rollout weight is $\lambda_R=0$ through the first 500 training epochs and $\lambda_R=1$ thereafter. Thus analytical solutions are used in training at $h=3$; the evaluation at other heights tests out-of-training-amplitude generalization, not independence from the analytical solution family.}

\subsubsection{Neural PDE}  
{The neural PDE model learns a grid-state tendency $F_\theta(U)\approx\partial_tU$ and advances it with RK4: $\widehat U^{n+1}=\mathcal R_{\Delta t}[F_\theta](\widehat U^n)$. After spatial discretization, this is a neural ODE on the discrete state~\cite{chen2019neuralordinarydifferentialequations}. {Training uses a trajectory target with 60 sampled states at relative times $0,\Delta t,\ldots,59\Delta t$, generated from the analytical soliton at $h=3$. The learned tendency is integrated during training, whereas the JAX rollout evaluates it four times per RK4 step.}}

\subsection{Implementation Framework}
The CPU solver uses NumPy~\cite{Harris2020NumPy}, while the
GPU solver and PolyNet use JAX~\cite{Bradbury2018JAX} to compile
the spatial operators and RK4 time stepping. The Emulator,
PINN, and NPDE are trained in PyTorch~\cite{Paszke2019PyTorch},
then evaluated in JAX using the trained weights. The Emulator
and PINN predict the next state directly; the NPDE predicts
the tendency, which is integrated using RK4. Thus, all GPU
methods use a common execution framework. TensorFlow is not
used in these experiments.

\section{1D Korteweg-De Vries Equation}
We recall that the 1D KdV equation is given by:
\begin{equation}
u_t+6uu_x+u_{xxx}=0,
\end{equation}
where $u(x,t)$ denotes the solution field. The balance between nonlinear advection and dispersion in this equation supports soliton solutions: localized waves that propagate at constant speed without changing shape. For real values of $x$, an exact single-soliton solution is:
\begin{equation}
u(x,t)=\frac{c}{2}\operatorname{sech}^{2}
\left[\frac{\sqrt{c}}{2}(x-ct-x_0)\right],
\label{eq:1d-KdV-soln}
\end{equation}
where $c>0$ is the propagation speed, $x_0$ is the initial peak position, and $h=c/2$ is the wave height. This known evolution provides a reference for assessing how well an integration method preserves the wave’s shape and motion.

For explicit methods such as RK4, resolving the dispersive term requires smaller time steps as the spatial grid is refined, increasing the cost of integration. Faithful soliton propagation also requires preserving both the wave profile and its position. Amplitude errors distort the wave height, whereas phase errors displace the wave from its expected location. Because soliton speed depends on amplitude, an amplitude error can also contribute to an accumulating position error. Controlling both errors is therefore necessary to predict the wave’s trajectory over long integration times.

{The simulations use the length-$40$ periodic grid $[-20,20)$, with $\Delta x=40/N$. The solitary-wave formula is exact on the real line; on the finite periodic grid it is used as a wrapped analytical reference, with finite-domain effects to be distinguished from numerical integration error~\cite{ZabuskyKruskal1965,DrazinJohnson1989}.}

The cost of integration becomes especially relevant when many independent trajectories must be computed. Ensembles of single-soliton simulations provide a controlled setting for examining how effectively each method uses parallel computation and how simulation speed changes as the number of trajectories increases. 

\begin{figure}[!ht]
\centering
\includegraphics[width=\linewidth]{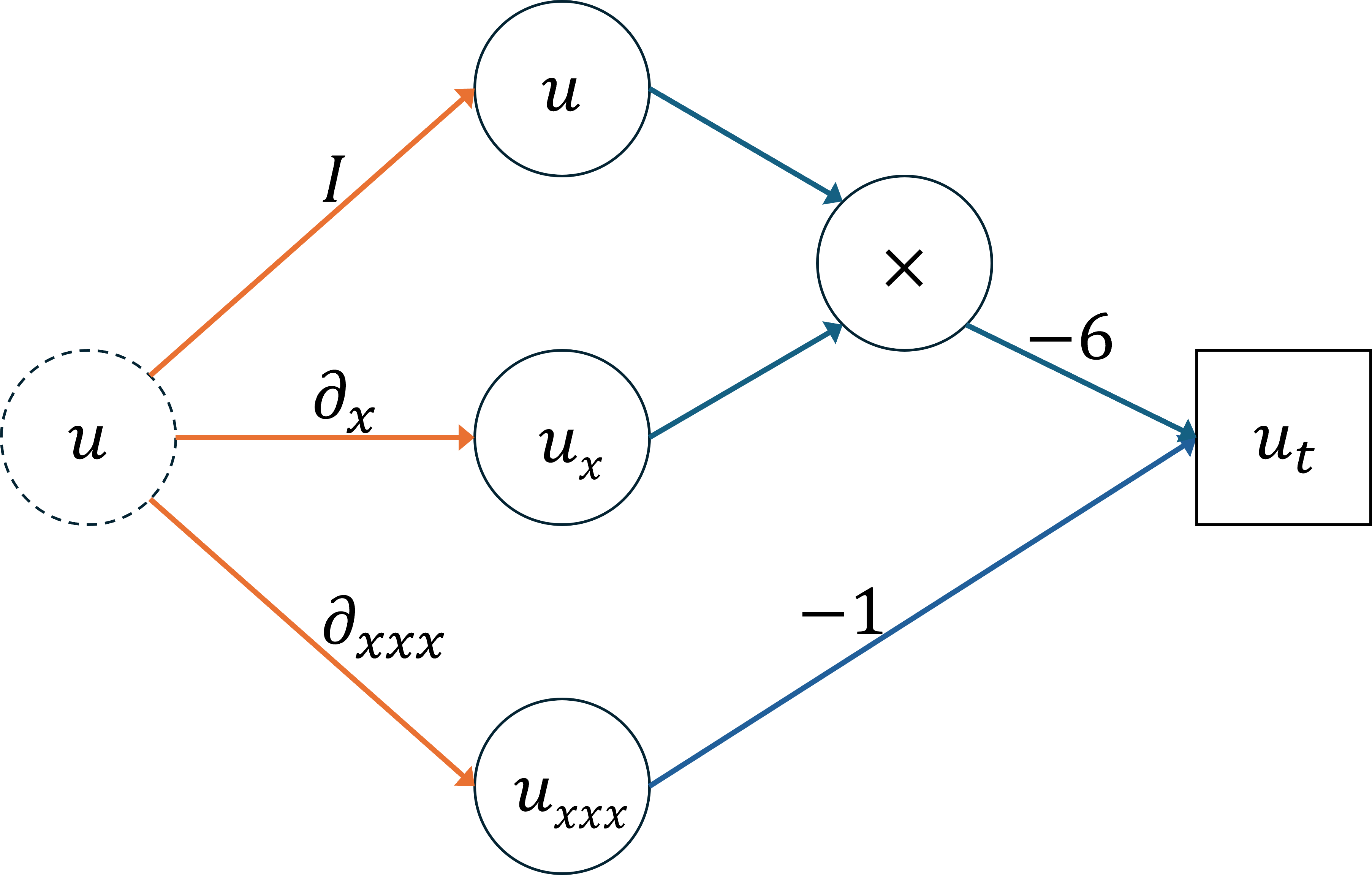}
\caption{{PolyNet graph for the semidiscrete KdV tendency, $F_N(U)=-6U\odot D_1U-D_3U$. Linear derivative branches and a componentwise product implement the prescribed physics. The four-stage recurrent RK4 time integrator is not shown.}}
\label{fig:kdv_polynet}
\end{figure}

What is noticeably unique about the methods emulating the solutions to this equation is that, of the six, the PolyNet is the only neural integration method that directly implements the KdV equation into its architecture. As seen in Figure \ref{fig:kdv_polynet}, rather than the neural architecture being a black-box method that may (PINN) or may not (Emulator, NPDE) be informed by the physics of the KdV soliton solutions, the PolyNet's architecture contains the deterministic equation as a feature of it, treating each neuron as a specific nonlinear operator rather than the general nonlinear activation functions of the classical neural methods (Emulator, PINN, NPDE).

{The spatial coordinate is used directly on a domain of length $L=40$; no $2\pi$-to-$40$ transformation is applied to the governing coefficients. Appendix~\ref{app:dispersive-timestep} gives the corresponding resolution-dependent time-step derivation.}

\subsection{Speed}

\begin{figure}[!ht]
\centering
\includegraphics[width=\linewidth]{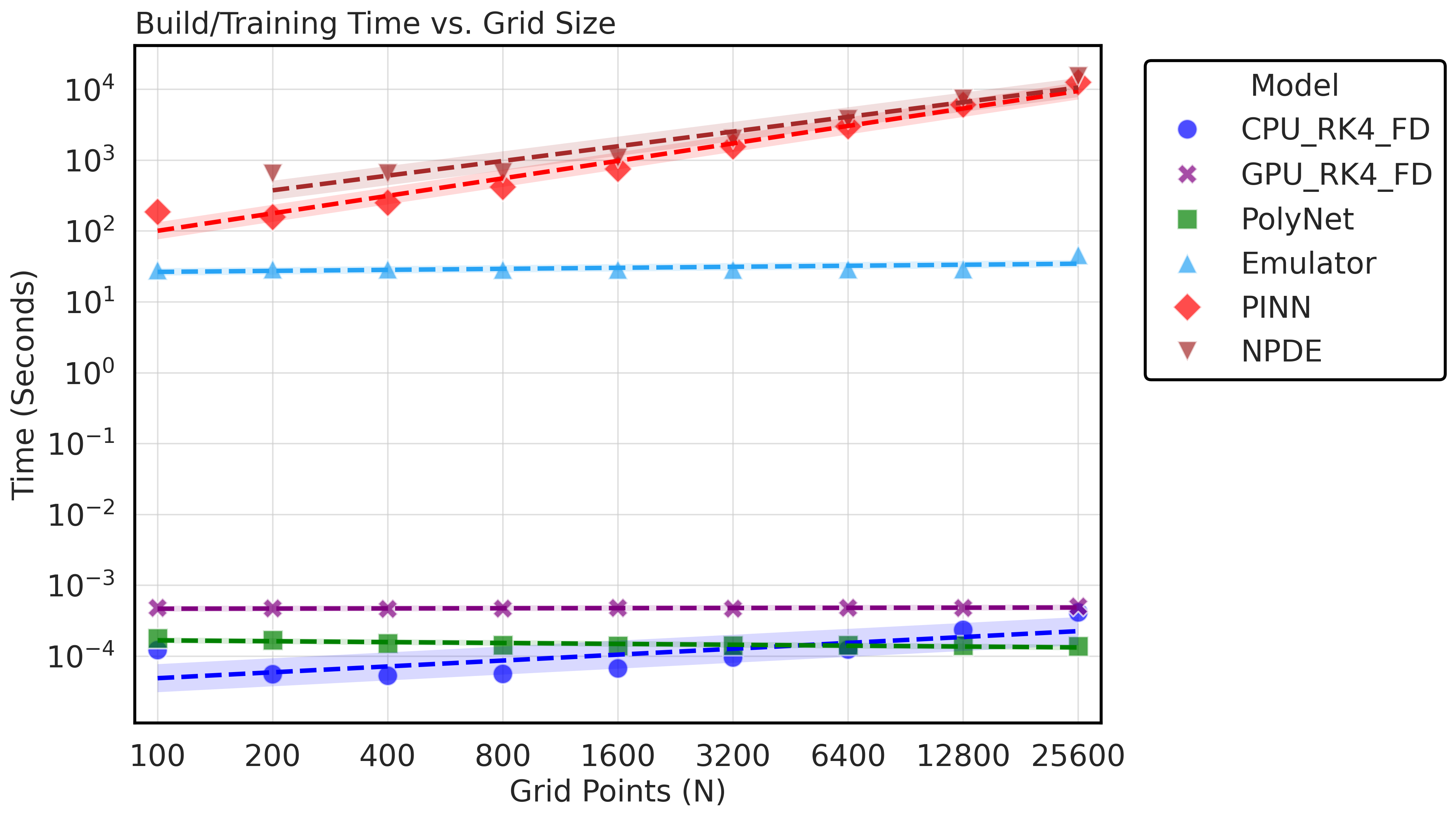}
\caption{{Training time for the three learned models and setup time for CPU finite-difference RK4, GPU finite-difference RK4, and PolyNet versus grid size $N$. The unsuccessful NPDE configuration at $N=100$ is omitted. Shading around log--log fits includes an imposed minimum display width and should not be interpreted as a confidence interval.}}
\label{fig:train}
\end{figure}

For each spatial resolution, we measured the computation time for 1000 blocks of 1000 time steps. Computation speed was calculated as $1000/T_{\mathrm{median}}$, where $T_{\mathrm{median}}$ is the median block integration time in milliseconds, yielding speed in steps/ms. Additionally, we recorded the training time for training-based models and compilation time for models that do not require training at each spatial resolution. Figure \ref{fig:train} shows that the training-based models add substantial runtime when compared to the non-trained methods. In particular, the PINN and the NPDE require longer training times than the Emulator. The PINN combines one-step analytical supervision, a spectral RK4 teacher, spatial-gradient matching, and an eight-step rollout loss. In contrast, the NPDE trains a learned tendency against a multistep trajectory through numerical integration. The  Emulator instead learns directly from input–output pairs. We note that PolyNet is a neural integration method that requires no training, so its setup cost consists only of code compilation. Its compilation time is comparable to that of the classical integration methods.

\begin{figure}[!ht]
\centering
\includegraphics[width=\linewidth]{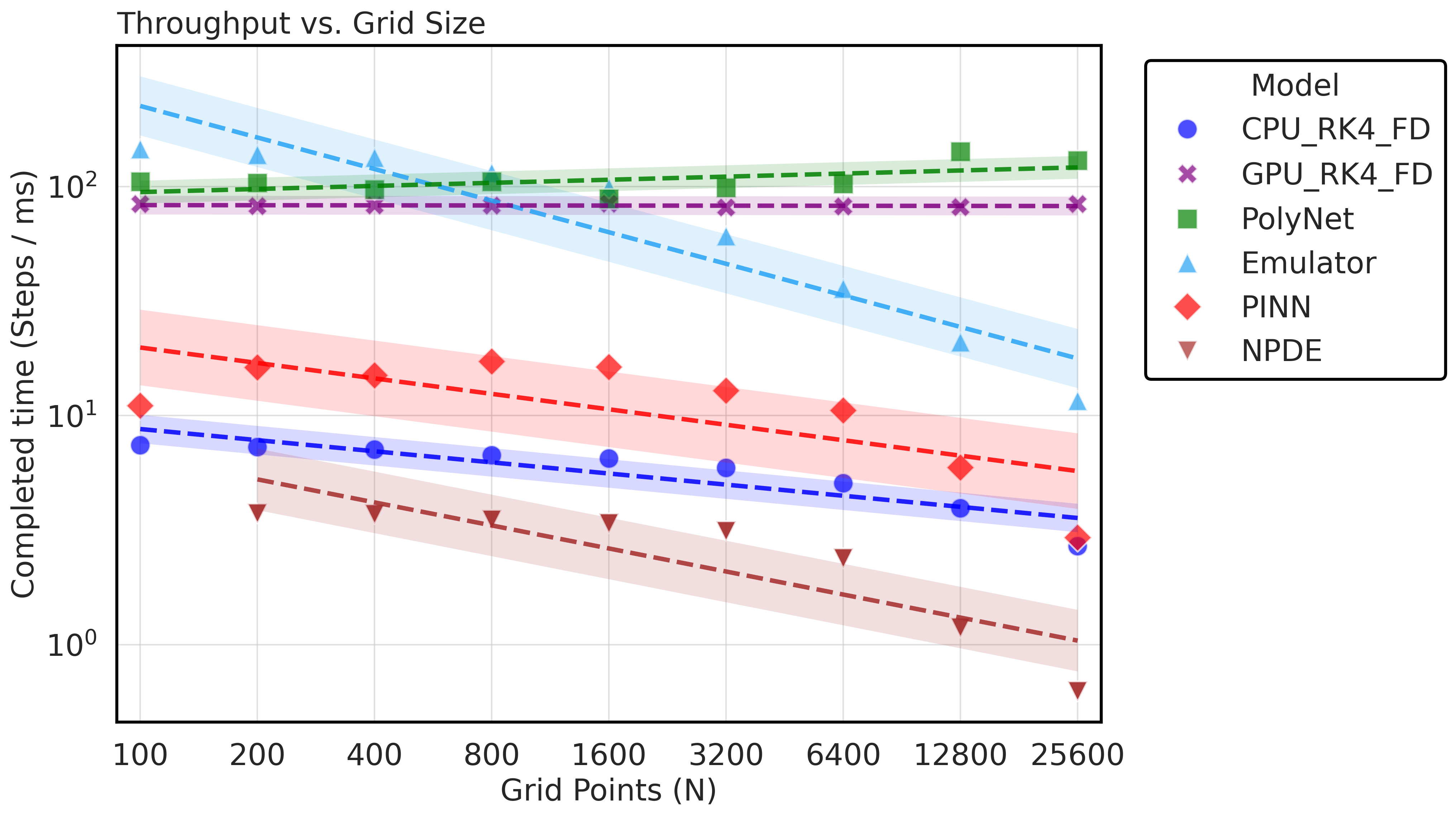}
\caption{{Time-step throughput versus grid size $N$. Each timed block advances $1000$ steps; throughput in steps/ms is $1000/T_{\mathrm{median}}$, where $T_{\mathrm{median}}$ is the median block duration in milliseconds. Shading around log--log fits includes an imposed minimum display width and is not a confidence interval.}}
\label{fig:speed}
\end{figure}

{Figure~\ref{fig:speed} compares measured steady-state throughput. The Emulator is fastest at coarse grids, while PolyNet and the conventional GPU finite-difference RK4 implementation maintain approximately level throughput across the tested range. These measurements demonstrate competitive GPU execution of the equation-derived graph; they do not isolate an intrinsic neural-hardware advantage or demonstrate training-induced sparsification.}

\subsection{Accuracy}

%For each grid resolution $N$, the maximum absolute phase error and maximum absolute amplitude error were collected for each model. This error was computed using the analytical solution discussed earlier. 

\begin{figure*}
\includegraphics[width=\linewidth]{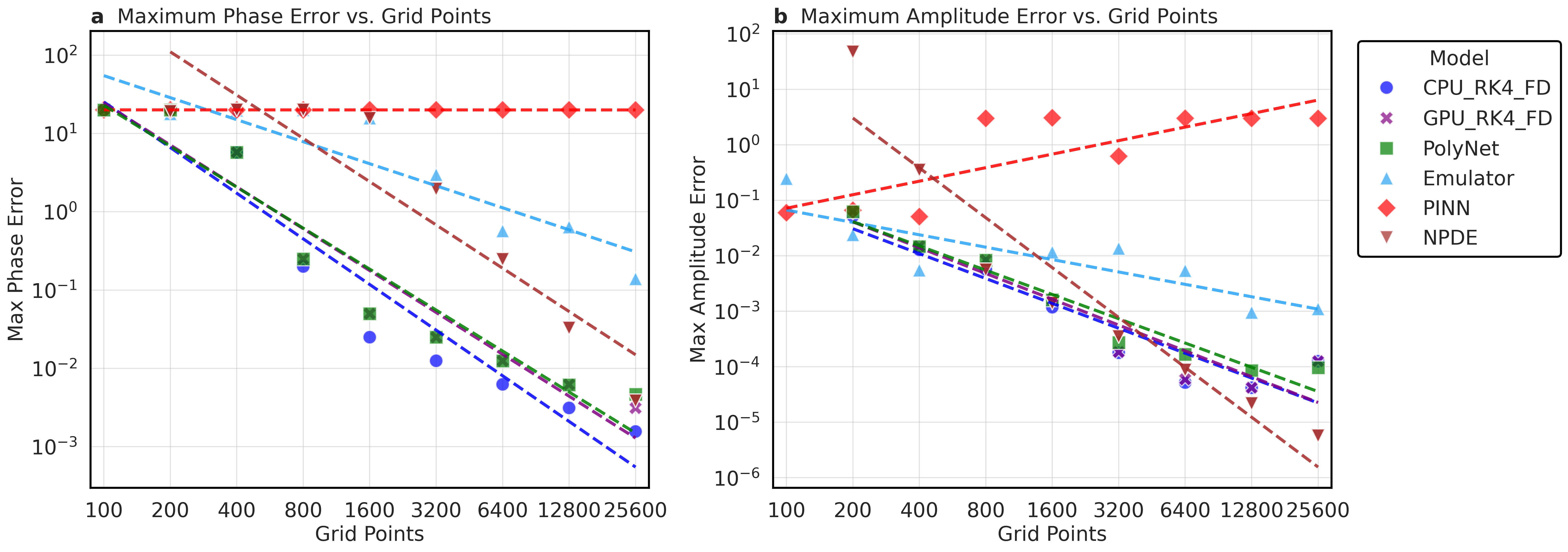}
\caption{{Maximum sampled absolute (a) phase and (b) amplitude errors against the analytical solitary-wave reference versus grid size $N$, over $10^6$ steps at the prescribed resolution-dependent dispersive time step. Unsuccessful GPU finite-difference RK4 configurations at $N=100$ and $N=200$ are omitted.}}
\label{fig:KdV-max-phase-amp-errors}
\end{figure*}

{At each resolution, we report maximum sampled phase and amplitude differences relative to the solitary-wave reference (Fig.~\ref{fig:KdV-max-phase-amp-errors}). PolyNet closely follows the corresponding CPU finite-difference RK4 errors. The trained models show more variable performance, and the amplitude-error ranking is not uniform across all grid sizes. The fixed $10^6$-step workload uses the resolution-dependent dispersive time step of Appendix~\ref{app:dispersive-timestep}; this is not a fixed-physical-time convergence experiment.}

\subsection{Ensemble}
\begin{figure}[!ht]
\centering
\includegraphics[width=\linewidth]{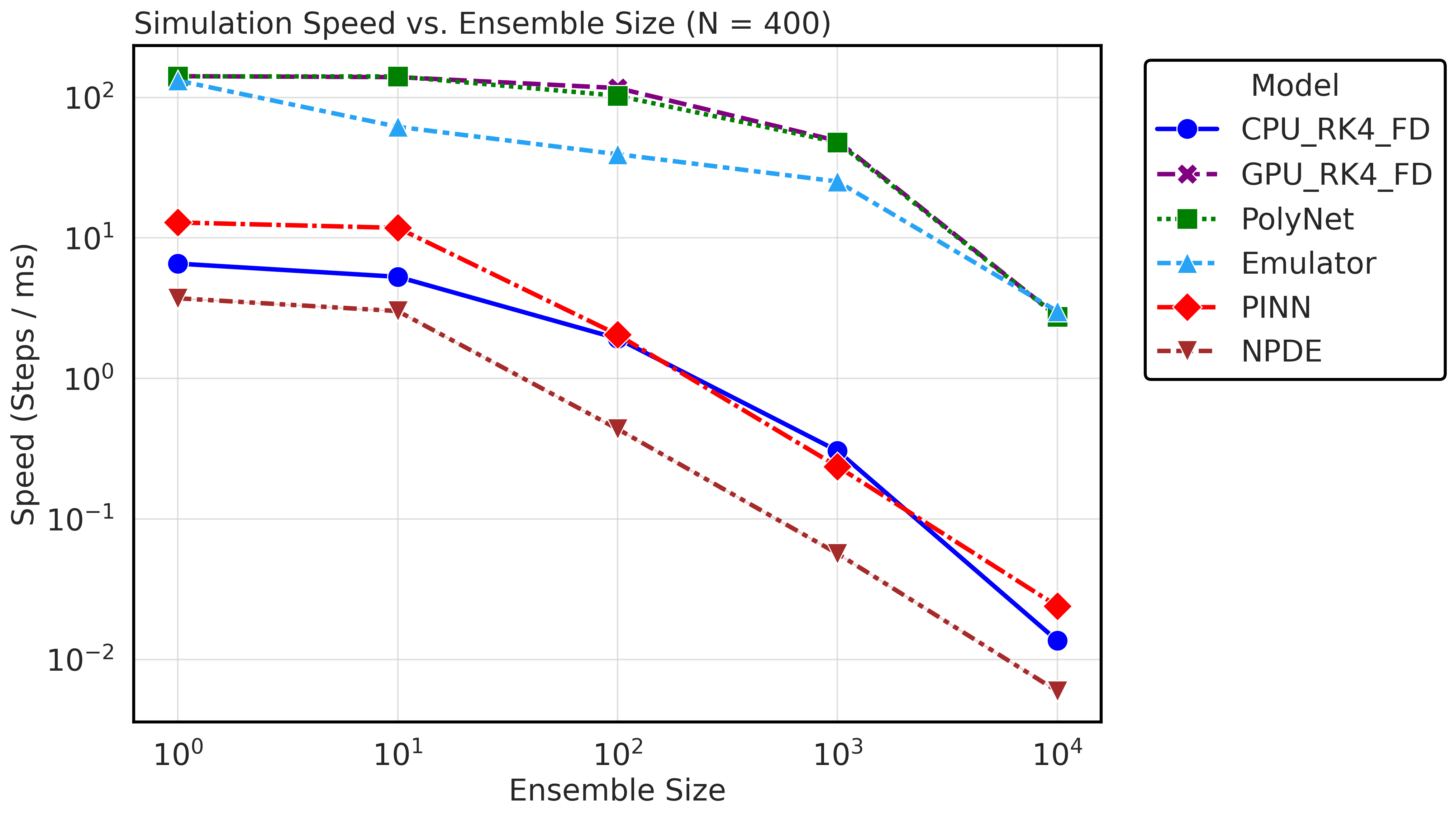}
\caption{{Median batch-step throughput (steps/ms) versus the number of independent soliton trajectories at $N=400$. Each batch step advances the entire ensemble; aggregate trajectory-step throughput is the plotted value multiplied by ensemble size.}}
\label{fig:ensemble}
\end{figure}

{Figure~\ref{fig:ensemble} compares the rate of advancing batched time steps at $N=400$ for $1$ to $10^4$ independent solitons. PolyNet and conventional GPU finite-difference RK4 closely track one another and lead over most tested batch sizes. This rate is distinct from aggregate trajectory-step throughput, which multiplies by ensemble size. The results support the utility of structured GPU computation for ensembles, without establishing a universal scaling law.}

\subsection{Generalization}
A known limitation of training-based methods is the poor ability to generalize beyond the domain of the training data \cite{kassai2024boosting,Sun2025,Wei2026}. In particular, these methods struggle with accurate extrapolation in physical modeling and often tend to converge to solutions encountered during training. What is unique about the PolyNet, however, is that expressing governing equations with known coefficients does not require any training. In the case when the coefficients are unknown, the network is trained on the solution to estimate the unknown coefficients and not to capture the features of the solution itself. This makes a PolyNet generalizable in the same way as the classical non-training-based methods. 

\begin{figure*}[!ht]
\includegraphics[width=\linewidth]{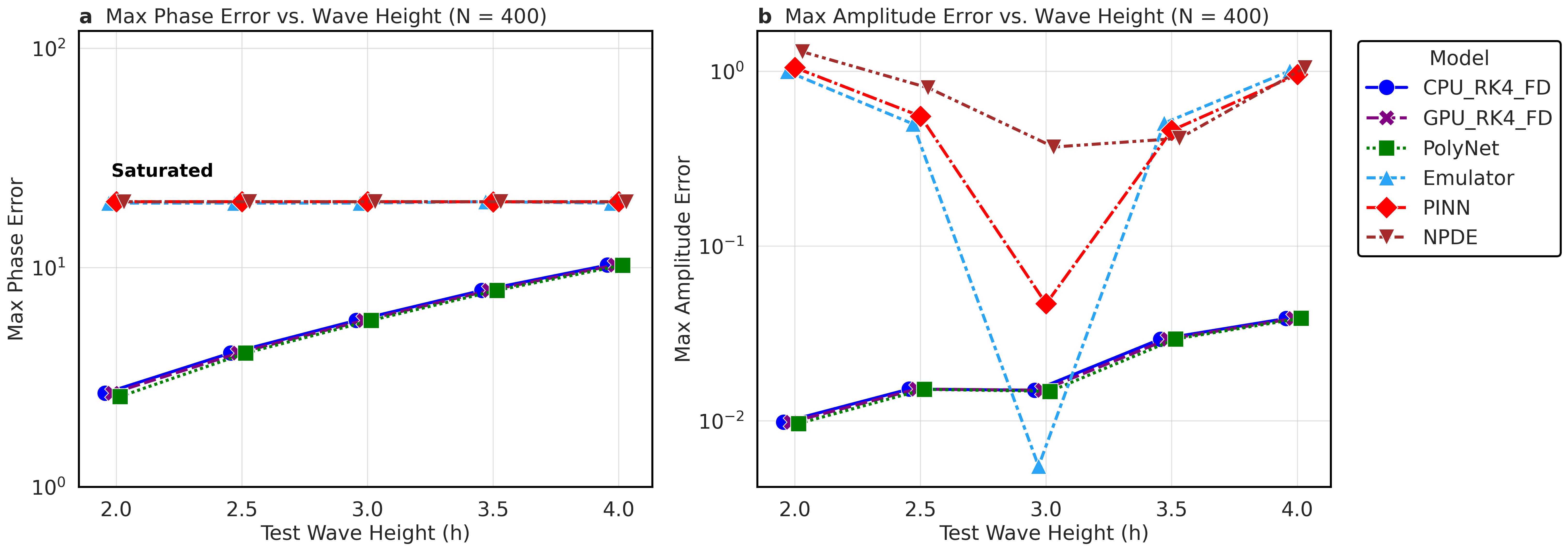}
\caption{{Maximum sampled absolute (a) phase and (b) amplitude errors versus initial soliton height $h$, for $N=400$ and $10^6$ integration steps. The neural models were trained at $h=3$ using analytical soliton targets; other heights were held out. Periodic phase separation is bounded by $L/2=20$, explaining the saturated phase-error curves.}}
\label{fig:generalization}
\end{figure*}

{Figure~\ref{fig:generalization} varies the initial height from $h=2$ to $h=4$ at fixed $N=400$, with $h=3$ the learned models' training height. PolyNet follows the prescribed numerical solvers, whereas the trained models exhibit substantial phase error even near the training height. Increasing errors for the numerical solvers are expected as taller solitons become narrower; Appendix~\ref{app:error-scaling} derives a conditional leading-order amplitude and speed error of $O(h^2\Delta x^2)$ and an accumulated unwrapped phase error of $O(h^2\Delta x^2t)$. The plotted periodic phase metric saturates at $L/2=20$.}

% \subsection{Implications} %Subsection for now

% \section{2D Kuramoto-Sivashinsky Equation}
% We recall that 
% \subsection{Speed} 

% \subsection{Accuracy}

% \subsection{Generalization}

\section{Summary \& Discussion}
{The six-way comparison separates equation-derived neural execution from learned state-map and tendency emulation. PolyNet reproduces the accuracy characteristics of its prescribed finite-difference RK4 scheme without training and achieves throughput competitive with a conventional GPU implementation. The Emulator is faster on coarse grids, while the GPU finite-difference method and PolyNet lead the ensemble comparison over most tested sizes. These results support equation-derived tensor computation; they do not establish that neural representation alone outperforms an equivalently compiled numerical graph.}

{PolyNet and the conventional finite-difference solvers retain coherent soliton propagation across the tested height range. The learned models exhibit larger phase errors, although their amplitude-error rankings vary with resolution and height. {The height dependence of numerical errors is consistent with the conditional leading-order scaling derived in Appendix~\ref{app:error-scaling}.}}

% In the realm of accuracy, the numerical methods (CPU\_RK4\_FD, GPU\_RK4\_FD) and the PolyNet noticeably outperformed the classic neural integrators (Emulator, PINN, NPDE), with this trend carrying on to the generalization of the methods to different initial conditions. Once again, the PolyNet separates itself from the pack as a neural integrator that maintains the speed of a simple neural integrator and the accuracy of classical numerical integration schemes, showcasing it's unique advantage to harness the power of both worlds in a unique way to other methods. 

{The practical advantage is architectural: the prescribed polynomial tendency and recurrent RK4 update execute within a compiled tensor graph without fitting a surrogate to the soliton trajectories. This preserves the chosen numerical method, including its discretization and stability limitations, while reserving learning for dynamics or coefficients that are not known.}

% From these results, it is clear that the PolyNet is a stand-out method for modeling dynamical systems when compared to competing methods, specifically due to its unique balance of throughput speed, scaling to large ensembles, accuracy to analytical solutions, and generalization to a broad range of initial conditions. This is likely attributed to a combination of factors that make the PolyNet unique. First and foremost, it is a neural architecture that takes an existing numerical scheme, such as RK4, and optimizes it for a GPU. This gives it the capacity to outperform CPU-based methods in terms of speed and large ensemble scaling due to the structure of a GPU. Further, because of its deterministic handling of both the classical numerical scheme as well as the polynomial dynamics that it models, in this case the KdV equation, it is not bound to the same training domain issues that other neural integrators are. Because of this, not only does it maintain the mathematical exactness to its CPU counterpart for the numerical schemes, handling accuracy up to float-point error, but it also handles generalization in a way that other neural integrators cannot. Finally, because of the implementation of the polynomial dynamics, the PolyNet only requires training if the coefficients of the dynamical system that we want to model are unknown, further lowering the computation cost of the method. 

The 1D setting is a limitation of the present study. {Future work should examine more complex multidimensional dynamics and compare PolyNet with classical GPU numerical methods and learned neural emulators.} Applications to dynamical Earth systems and large-scale chaotic dynamics would also help assess its performance beyond the present setting. In particular, applying PolyNet to the shallow-water equations would provide a test of its accuracy, throughput, and scalability for these dynamics.

% As a direct result of these findings, future work should focus on implementing this method into mode complicated dynamics and comparing their results. Specifically, a limitation of this work is the one-dimensional nature of the problem we've addressed. Because of this, a multi-dimensional problem, such as modeling the Kuramoto-Sivashinsky equation in two-dimensions, is of great interest, specifically in comparing the PolyNet to the similar models, such as the simple neural integrator (Emulator) as well as the numerical integrators (CPU\_RK4\_FD, GPU\_RK4\_FD). There is also significant interest in real-world modeling using this method, and if it may prove to be more effective in modeling dynamical Earth systems and large-scale chaotic dynamics. An example of such work would be attempting to use PolyNets to model the Shallow Water equations, highlighting its efficacy at modeling these large-scale real-world dynamics. Future work should focus on problems such as these, as they would provide a true test to the strength of the PolyNet method. 

\section{Conclusions}

{For the known polynomial semi-discrete KdV dynamics tested here, a training-free PolyNet provides a neural tensor-graph realization of prescribed finite-difference RK4 integration. Its numerical fidelity is comparable to that of the corresponding classical scheme, and its GPU throughput is competitive over the tested grids and ensembles. This demonstrates that neural emulation need not require learning a surrogate when the governing polynomial vector field is available; it does not eliminate the value of learned models for unknown, unresolved, or nonpolynomial dynamics.  In ongoing work, we apply this approach to 2D turbulence and shallow water equations.}

\section*{Acknowledgments}

This material is based in part upon work supported by the Bangkok Bank, Schmidt Sciences, Community Jameel for Jameel Observatory CREWSnet, and by the MIT Climate Grand Challenges. Luka Todorovic acknowledges the MIT Summer Research Program (MSRP) for sponsoring this work. Kaushal Gianchandani’s research was supported by an appointment to the NASA Postdoctoral Program at NASA Goddard Institute for Space Studies from the NASA Astrobiology Program, administered by Oak Ridge Associated Universities, under contract with NASA.  The authors acknowledge the use of ChatGPT (OpenAI) and Grammarly to assist with drafting, revising, and editing the manuscript. The authors retain full responsibility for the content, accuracy, and integrity of the work, including its ideas, analyses, interpretations, and conclusions.

% Use \appendix only once if the manuscript has other appendices.
%
% Suggested replacement sentence in Methods:
% At each spatial resolution, all methods use the prescribed time step
% $\Delta t\simeq(\Delta x)^3/6$, with $\Delta x=40/N$, corresponding to
% $\Delta t\lVert D_3\rVert_\infty\simeq0.5$ for the third-derivative
% stencil used here (Appendix~\ref{app:dispersive-timestep}).
%
% Source basis: kdv_soliton-main/generalized_exp/gen_exp_polynet.py,
% lines 18--24 (domain and time-step table), 33--64 (stencils and RK4),
% and 85--86 (periodic spatial grid). The norm and spectral analysis below
% are derived from that implementation. The source does not document the
% original rationale for the prefactor in its base time-step table.

\appendix
\section{Dispersive time-step selection and linear stability}
\label{app:dispersive-timestep}
{The operator-norm and RK4 stability calculation below follows standard finite-difference and Runge--Kutta stability analysis~\cite{Butcher2008,HundsdorferVerwer2003,LeVeque2007}. The specific stencil and time-step prefactor are those used in the archived implementation, rather than values taken from these references.}
\subsection{Dispersive scaling}

For the nondimensional Korteweg--de Vries equation,
\begin{equation}
    u_t+6uu_x+u_{xxx}=0,
    \label{eq:app-kdv}
\end{equation}
we use a uniform periodic grid on $[-L/2,L/2)$ with $L=40$,
\begin{equation}
    x_j=-\frac{L}{2}+j\Delta x,
    \qquad \Delta x=\frac{L}{N}=\frac{40}{N},
    \label{eq:app-grid}
\end{equation}
where $j=0,\ldots,N-1$. In the linear dispersive equation
$u_t+u_{xxx}=0$, substituting a Fourier mode
$u=\widehat u\exp[\mathrm{i}(k_mx-\omega_mt)]$ gives
\begin{equation}
    k_m=\frac{2\pi m}{L},
    \qquad \omega_m=-k_m^3.
    \label{eq:app-dispersion}
\end{equation}
The highest resolved wavenumbers scale as $(\Delta x)^{-1}$, so their
characteristic dispersive time scales as $(\Delta x)^3$. Maintaining
a fixed dimensionless time increment for these modes therefore requires
$\Delta t\propto(\Delta x)^3$. The admissible prefactor depends on the
spatial discretization and time integrator, rather than on the domain
length alone.

\subsection{Implemented time step and its normalization}

The implementation prescribes a resolution-dependent table of base time
steps satisfying $\Delta t_{\mathrm{base}}\simeq(\Delta x)^3/3$ and
multiplies each entry by $0.5$. Consequently, the time step used at each
resolution is
\begin{equation}
    \Delta t=0.5\,\Delta t_{\mathrm{base}}
    \simeq\frac{(\Delta x)^3}{6}
    =\frac{1}{6}\left(\frac{40}{N}\right)^3.
    \label{eq:app-timestep}
\end{equation}
Here $\simeq$ accounts for rounding of the tabulated values. The time
step is fixed within each run, not adjusted adaptively as the solution
evolves. Doubling $N$ reduces $\Delta t$ by approximately a factor of
eight.

The implemented centered third-derivative operator is
\begin{equation}
    (D_3U)_j=
    \frac{U_{j+2}-2U_{j+1}+2U_{j-1}-U_{j-2}}
         {2(\Delta x)^3},
    \label{eq:app-d3}
\end{equation}
with indices taken modulo $N$. Its induced matrix infinity norm is the
maximum absolute row sum,
\begin{equation}
    \lVert D_3\rVert_\infty
    =\frac{1+2+2+1}{2(\Delta x)^3}
    =\frac{3}{(\Delta x)^3}.
    \label{eq:app-d3-norm}
\end{equation}
Thus two explicit normalizations of the prescribed time step are
\begin{align}
    C_{\mathrm{disp}}
    &\equiv\frac{\Delta t}{(\Delta x)^3}
    \simeq\frac{1}{6},
    \label{eq:app-cdisp}\\
    C_{\mathrm{op}}
    &\equiv\Delta t\lVert D_3\rVert_\infty
    =3C_{\mathrm{disp}}
    \simeq0.5.
    \label{eq:app-cop}
\end{align}
The values $1/6$ and $0.5$ therefore describe the same time-step choice
under different normalizations. The operator-norm identity characterizes
the prescribed table; it neither determines the prefactor from the
continuous equation nor identifies a sharp stability limit.

\subsection{RK4 stability for the linear dispersive operator}

For a discrete Fourier mode $U_j=\exp(\mathrm{i}j\theta)$,
with $\theta=2\pi m/N$, the symbol of Eq.~\eqref{eq:app-d3} is
\begin{equation}
    \widehat D_3(\theta)
    =-\frac{2\mathrm{i}\sin\theta(1-\cos\theta)}{(\Delta x)^3}.
    \label{eq:app-symbol}
\end{equation}
The eigenvalues $\lambda_m$ of the linear evolution operator $-D_3$
are purely imaginary. Maximizing
$2|\sin\theta(1-\cos\theta)|$ over $\theta$ gives $3\sqrt{3}/2$,
attained at $\cos\theta=-1/2$, and hence
\begin{equation}
    \max_m|\lambda_m|
    \leq\frac{3\sqrt{3}}{2(\Delta x)^3}.
    \label{eq:app-spectrum}
\end{equation}
The inequality allows for the fact that a finite grid need not contain
the maximizing wavenumber.

The classical fourth-order Runge--Kutta method has stability polynomial
\begin{equation}
    R(z)=1+z+\frac{z^2}{2}+\frac{z^3}{6}+\frac{z^4}{24}.
    \label{eq:app-rk4}
\end{equation}
For real $y$, direct substitution yields
\begin{equation}
    |R(\mathrm{i}y)|^2
    =1-\frac{y^6}{72}+\frac{y^8}{576}
    =1+\frac{y^6(y^2-8)}{576}.
    \label{eq:app-rk4-modulus}
\end{equation}
Thus $|R(\mathrm{i}y)|\leq1$ for $|y|\leq2\sqrt{2}$. Because the
periodic difference operator is diagonalized by the discrete Fourier
transform, combining this interval with Eq.~\eqref{eq:app-spectrum}
gives the sufficient bound
\begin{equation}
    \Delta t\leq
    \frac{4\sqrt{2}}{3\sqrt{3}}(\Delta x)^3
    \approx1.089\,(\Delta x)^3
    \label{eq:app-rk4-bound}
\end{equation}
for linear $\ell^2$ stability of the isolated dispersive problem.
For the nominal choice $\Delta t=(\Delta x)^3/6$,
\begin{equation}
    \Delta t\max_m|\lambda_m|
    \leq\frac{\sqrt{3}}{4}
    \approx0.433<2\sqrt{2},
    \label{eq:app-margin}
\end{equation}
so the prescribed steps lie well within this linear dispersive
stability bound. This estimate applies to the stated periodic
finite-difference operator and RK4; it does not by itself establish
stability of the full nonlinear KdV calculation or of a learned
evolution map.

% Insert after Appendix A, before the bibliography.
% Do not repeat \appendix: the next \section becomes Appendix B.
% This is a leading-order, near-soliton scaling argument, not a fit to Figure 6.

\section{Amplitude and phase error scaling}
\label{app:error-scaling}
{The soliton parameterization follows the classical KdV literature~\cite{ZabuskyKruskal1965,DrazinJohnson1989}; the Taylor expansions use standard centered-difference truncation analysis~\cite{LeVeque2007,Fornberg1988}. The conditional height-scaling argument is derived here and should not be attributed to those references as a proved error law.}

Figure~\ref{fig:generalization} varies the initial soliton height while
holding the spatial grid, time step, and number of integration steps
fixed. Here we derive leading-order error estimates for the common
centered finite-difference discretization used by the classical
integrators and PolyNet. The estimates describe the regime in which
spatial truncation error dominates and the numerical solution remains
a small perturbation of a coherently translating soliton. They are not
unconditional error bounds for arbitrary integration times or for the
trained neural models.

\subsection{Height-dependent spatial truncation error}

Writing the soliton in terms of its height $h=c/2$ gives
\begin{equation}
 U_h(x,t)=h\,\operatorname{sech}^{2}
 \left[\kappa(x-2ht-x_0)\right],
 \qquad \kappa=\sqrt{h/2}.
 \label{eq:app-soliton-height}
\end{equation}
Its speed is $c=2h$, and its characteristic width is
$\ell=\kappa^{-1}=\sqrt{2/h}$. Thus a taller soliton is also narrower
and is resolved by fewer grid points when $\Delta x$ is held fixed.
With $f(\eta)=\operatorname{sech}^{2}\eta$,
\begin{equation}
 \partial_x^m U_h=h\kappa^m f^{(m)}
 \left[\kappa(x-2ht-x_0)\right].
\end{equation}
Consequently, the dimensionless parameter controlling the leading
spatial error is
\begin{equation}
 \epsilon_x=(\kappa\Delta x)^2
 =\frac{h(\Delta x)^2}{2}.
 \label{eq:app-small-parameter}
\end{equation}

For smooth resolved profiles, Taylor expansion of the implemented
centered stencils yields
\begin{align}
 D_1U &= U_x+\frac{(\Delta x)^2}{6}U_{xxx}
          +O((\Delta x)^4),\\
 D_3U &= U_{xxx}+\frac{(\Delta x)^2}{4}U_{xxxxx}
          +O((\Delta x)^4).
\end{align}
The discrete tendency $F_{\Delta x}[U]=-6U D_1U-D_3U$
therefore differs from $F[U]=-6UU_x-U_{xxx}$ by
\begin{align}
 F_{\Delta x}[U]-F[U]
 ={}&-(\Delta x)^2
       \left(UU_{xxx}+\frac14 U_{xxxxx}\right)
       +O((\Delta x)^4).
 \label{eq:app-tendency-defect}
\end{align}
For $U=U_h$, each leading physical tendency scales as $h^{5/2}$,
whereas the leading defect scales as $(\Delta x)^2h^{7/2}$.
The relative spatial perturbation is therefore
$O(h(\Delta x)^2)$, equivalently $O(\epsilon_x)$.

\subsection{Amplitude bias and accumulating phase error}

The soliton's characteristic evolution time is
$\tau_s\sim\ell/c\sim h^{-3/2}$. A tendency defect of size
$O((\Delta x)^2h^{7/2})$ acting over this time has magnitude
$O((\Delta x)^2h^2)$. In a stable near-soliton regime with a
bounded shape adjustment, this motivates the amplitude-error estimate
\begin{equation}
 E_A\sim K_A h^2(\Delta x)^2.
 \label{eq:app-amplitude-scaling}
\end{equation}
Likewise, a relative speed perturbation of order $\epsilon_x$ gives
\begin{equation}
 |\delta c|\sim K_\phi h^2(\Delta x)^2.
 \label{eq:app-speed-scaling}
\end{equation}
The coefficients $K_A$ and $K_\phi$ are not determined by this scaling
argument. They depend on the numerical scheme and the response of the
soliton to its perturbation; the estimates assume that this response
has comparable leading-order coefficients over the height range.

Let $X_{\mathrm{num}}(t)$ be the unwrapped numerical peak position
and define $\delta X(t)=X_{\mathrm{num}}(t)-(x_0+2ht)$.
For an initially aligned wave,
\begin{equation}
 \delta X(t)=\int_0^t\delta c(s)\,ds.
 \label{eq:app-phase-accumulation}
\end{equation}
A persistent, approximately constant speed bias thus produces
\begin{equation}
 E_\phi^{\mathrm{unwrapped}}(t)
 \sim K_\phi h^2(\Delta x)^2t.
 \label{eq:app-phase-scaling}
\end{equation}
Hence a small bounded amplitude or shape error can coexist with a
phase error that grows with integration time. No corresponding
linear-in-time growth of amplitude error is implied by
Eq.~\eqref{eq:app-amplitude-scaling}.

For a perturbation confined to the exact soliton family, the
amplitude--speed relation additionally gives $\delta c=2\delta h$.
Its amplitude-induced phase contribution is then
\begin{equation}
 \delta X_A(t)=2\int_0^t\delta h(s)\,ds.
 \label{eq:app-amplitude-phase-coupling}
\end{equation}
A constant signed amplitude bias $\delta h=a$ produces
$|\delta X_A(t)|=2|a|t$. If instead the signed amplitude bias drifts
as $\delta h(t)=\alpha t$, its phase contribution is
$\delta X_A(t)=\alpha t^2$. These conditional relations do not
identify the full numerical speed error with twice the measured peak
error: discrete dispersion and non-soliton shape changes can also
contribute to phase drift.

\subsection{Interpretation of the height sweep and periodic saturation}

For Figure~\ref{fig:generalization}, $L=40$ and $N=400$, so
$\Delta x=0.1$. The common prescribed time step is approximately
$1.666665\times10^{-4}$; $10^6$ steps therefore correspond to
$T\simeq166.67$ for every initial height in this figure.
In the regime described above, the leading-order estimates at this
fixed horizon are
\begin{equation}
 \boxed{
 E_A(h)\propto h^2,
 \qquad E_\phi^{\mathrm{unwrapped}}(h;T)\propto h^2.
 }
 \label{eq:app-height-scaling}
\end{equation}
Thus doubling the initial height suggests approximately four times
as much amplitude error and unsaturated phase error, rather than
height-independent accuracy. This is an expected discretization
trend, not a fitted law or a statement that each measured point must
follow an exact power law. Because the classical finite-difference
integrators and PolyNet evaluate the same spatial tendency and RK4
scheme, the same leading-order argument applies to each of them.

The phase diagnostic uses the shortest distance between peaks on the
periodic domain. In terms of unwrapped displacement, its ideal
continuous-time form is
\begin{equation}
 e_\phi(t)=\min_{m\in\mathbb Z}|\delta X(t)-mL|
 \leq\frac L2=20.
 \label{eq:app-periodic-phase}
\end{equation}
For constant speed bias, exact initial alignment, and continuous-time
evaluation, the running maximum is
\begin{equation}
 \max_{0\leq t\leq T}e_\phi(t)
 =\min\left(|\delta c|T,\frac L2\right).
 \label{eq:app-phase-saturation}
\end{equation}
The code evaluates errors at sampled times, so its maximum can
approach this ceiling without attaining it exactly. A speed bias of
about $20/T\simeq0.12$ is sufficient to reach a half-domain
separation within the Figure~\ref{fig:generalization} horizon under
this constant-bias model. Individual wrapped errors need not plateau:
it is their running maximum that saturates. The plateau labeled
``Saturated'' therefore reflects the phase metric's ceiling and does
not establish that the underlying unwrapped phase error has ceased
to grow.

These estimates explain the expected height dependence of the
prescribed numerical methods and the periodic phase ceiling. They
do not determine the errors of the trained Emulator, PINN, or NPDE,
whose learned representations may introduce additional amplitude,
shape, and speed biases. In particular, a small amplitude error near
the training height $h=3$ does not ensure accurate phase over a long
rollout.

\bibliography{ref}% Produces the bibliography via BibTeX.

\end{document}